# Vehicles Detection Based on Background Modeling

Mohamed Shehata, Reda Abo-Al-Ez, Farid Zaghlool, and Mohamed Taha
*Department of Systems & Computers Engineering, Faculty of Engineering, Al-Azhar University, Egypt*

*Abstract*

  Background image subtraction algorithm is a common approach which detects moving objects in a video sequence by finding the significant difference between the video frames and the static background model. This paper presents a developed system which achieves vehicle detection by using background image subtraction algorithm based on blocks followed by deep learning data validation algorithm. The main idea is to segment the image into equal size blocks, to model the static reference background image (SRBI), by calculating the variance between each block pixels and each counterpart block pixels in the adjacent frame, the system implemented into four different methods: Absolute Difference, Image Entropy, Exclusive OR (XOR) and Discrete Cosine Transform (DCT). The experimental results showed that the DCT method has the highest vehicle detection accuracy.

## I. INTRODUCTION

   The increasing number of vehicles makes a lot of pressure on roads capacity and infrastructure. Accordingly the traffic management is too difficult and the traffic hazards increases at a high rate, moreover it causes a huge loss of life and property due to the road accidents. Vehicle detection systems play an important role to reduce the harmful side effects of traffic, as a part of many traffic applications such as road traffic control, traffic response system, traffic signal controller, lane departure warning system, automatic vehicle accident detection and automatic traffic density estimation [1]. Moving vehicles detection is usually carried out by using background image subtraction techniques. The key behind this kind of techniques is to first build a background model from a sequence of images in order to find the moving vehicles from the difference between that background estimation and the current frame [2]. Natural scenes in real life are usually composed of several dynamic entities, which make the background extraction problem more complex [3]. Such as swaying vegetation, fluctuating water, flickering monitors, ascending escalators, etc., or gradual or sudden changes in illumination also changes in the background geometry such as parked cars, and so on. We developed a system which is robust and generic enough to handle the detecting vehicles in the complexities of most natural dynamic scenes. The performance of the system is thus highly dependent on the accuracy of the background image subtraction.

   The paper is organized as follows: Section II provides a briefing of related works. Overview of the overall developed systems is explained in Section III. Section IV deals with experimental results. And the paper is concluded with Section V.

## II. RELATED WORKS

   The commonly used strategy in many background image subtraction systems, are median filter method and mixed-Gaussian methods, but they have some shortcomings such as imprecise initial background, maintaining and updating difficulties of background model. Some of the related works are illustrated as follows:

   Yan Zhang [4], presented a block-wise mixture models to model a background image and detect moving objects from a video frames. Features of each block are extracted by the method of Integral Image. The models are created and updated by the scheme of mixture models.

   Dorra Riahi [5], developed a system to background image subtraction based on rectangular regions. The general principle is to successively divide the image into blocks and detect foreground pixels based on the color histogram and the variance between pixels of the blocks. Then, the classic Gaussian Mixture background image subtraction method is applied to refine the detected foreground.

   Jing-Ming Guo [6], presents a cascaded scheme with block-based and pixel-based codebooks for background image subtraction. Then codebook is mainly used to compress information. In the block-based stage, 12 intensity values are employed to represent a block. The algorithm extends the concept of the Block Truncation Coding; he also presented a color model and a match function which can classify an input pixel as shadow, highlight, background, or foreground.

   Igor Lipovac [7], addressed three practical issues which arose while applying background modeling for detecting vehicle presence in urban intersection video. First, background model based on a Histogram of Oriented Gradients (HOG) cells has achieved nearly the same detection accuracy as texture decomposition while requiring much less computational resources. Second, the HOG background model paired with the integral gradient images. Third, he proposed a two-stage modification to the widely used Gaussian update approach, in order to better approximate the classic running average in the desired time interval.

### III. VEHICLE DETECTION DEVELOPED SYSTEM

Moving vehicles detection is using background image subtraction to complete extraction of moving vehicles, and foreground is obtained by subtracting the background model from the current video frame. The key in successfully background image subtraction depend on the success of background modeling process which produces the SRBI, and because the background often dynamically changes with the light, movement and target in and out the scene. It needs an efficient strategy to maintain and update the background model. The vehicle detection flowchart is illustrated in Fig.1.

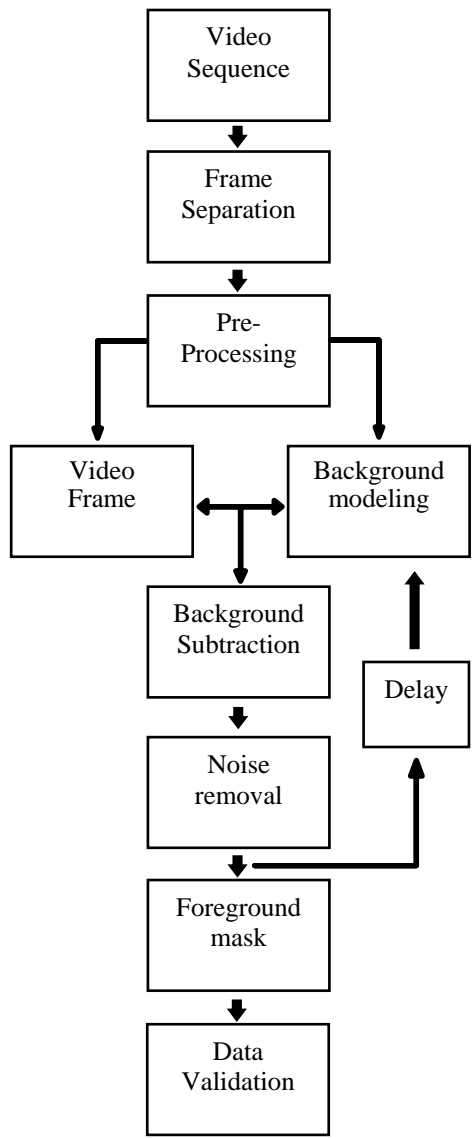

**Fig 1: Vehicle Detection Process**

After video separated into frames, an image pre-processing step is required to achieve cancellation of natural noise like, rain, dust and fog in open air, in addition to handle some defects caused by the camera like blurred and noise. Then a background modeling algorithm has applied to extract the SRBI. Where the first frame is segmented into equal rectangular blocks; the block ratio equal the image ratio, and the block size calculated automatically, depending on the traffic density and congestion in the image, which can be obtained by comparing the difference between the image entropy of the first frame and the image entropy of the followed frame, if it was a big difference obtained, that mean there is a congestion in the image, which in turn summons small block size, and vice versa.

The entropy of an image is mathematically defined as the following equation:

$$H = \sum_{i=0}^{n-1} p_i \log_b p_i$$

Where (H) is image entropy, (n) is the number of gray levels (256 for 8-bit images), ($P_i$) is the probability of a pixel having grey level (i), and (b) is the base of the logarithm function[8].

We calculated the variance between each block pixels and the counterpart block pixels, in the adjacent frame, then compare the output result to a certain threshold, for each block if comparative value is below the threshold value, we save this block in a buffer as a true piece of the SRBI, but if not, we have to repeat this process, with the next adjacent frames, until the required SRBI completely model successfully, by grouping the successful obtained blocks together, as shown in Fig. 2. Then we repeat the whole SRBI modeling process again to update the SRBI, which is necessary to recover any changes in background like vehicles get barked or leave barking.

We used four methods to calculate the blocks pixels variance as following:

- First method depends on absolute difference by direct subtraction for each two counterpart block pixels in adjacent frames.
- Second method depends on comparing the image entropy for each two counterpart block pixels in adjacent frames.
- Third method depends on applying XOR function [9], to detect changes in the each two counterpart block pixels in adjacent frames.
- Fourth method depends on features of each two counterpart block pixels in adjacent frames, are extracted by applying DCT, as each block pixels are transformed from the spatial domain to the frequency domain, afterwards saving the result in a weight vector. The effective features are then selected using a Zigzag technique [10], and the similarity of the two block pixels, is quickly calculated by comparing their features.

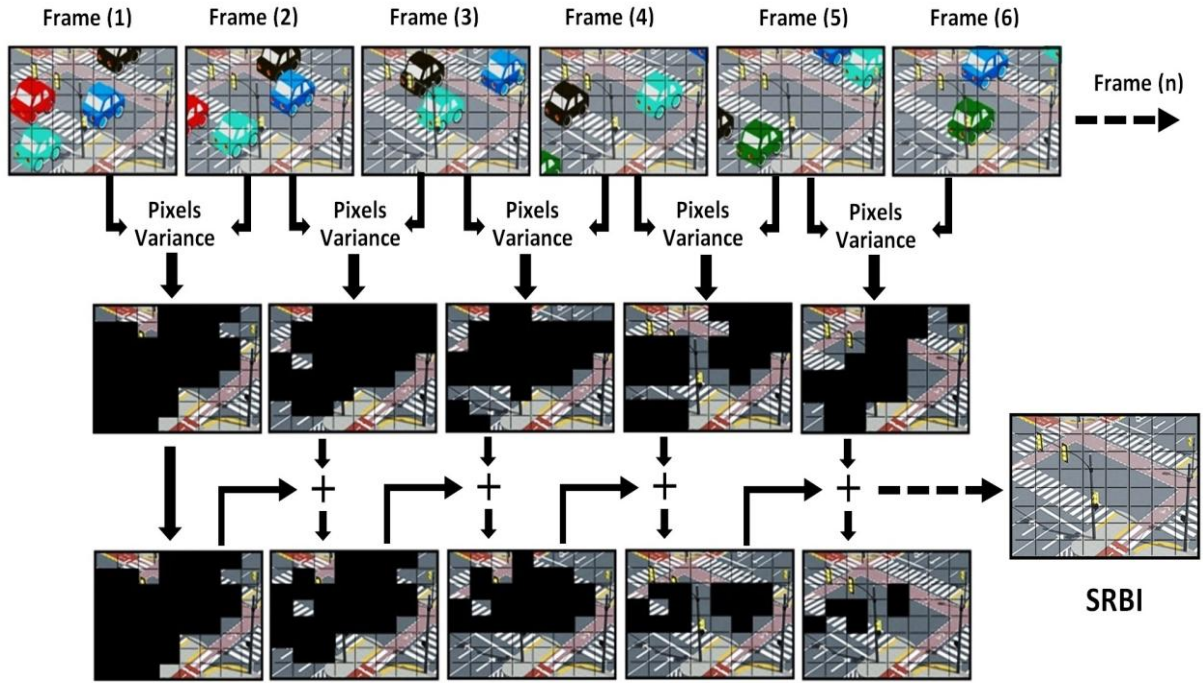

**Fig 2: Background Modelling Example**

After modeling the SRBI we applied a detection algorithm, to detect the moving objects from the current video frame image. We used XOR function to detect changes between two images, since pixels which didn't change output 0 and pixels which did change result in 1. In Fig. 3., we can see the SRBI and the video frame image contains a moved object, whereas the stationary parts almost disappeared in the XOR result. Due to the effects of noise, we can still see some pixels in the stationary parts.

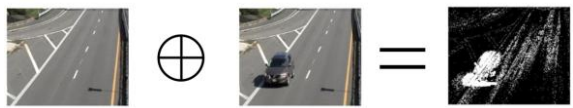

**Fig 3: Background image subtraction Using XOR**

A noise cancellation filter is needed to improve the candidate foreground mask based on information obtained from background image subtraction, as we eliminates noise that do not correspond to actual moving objects, by applying median filter [11], as shown in Fig 4.

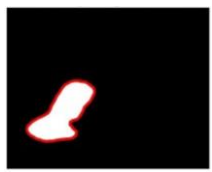

**Fig 4: Foreground Mask**

The output of the last process is a foreground mask which represents by a binary image where background pixels are labeled as 0 and any foreground objects are labeled as 1. Then we apply this mask to the video frame to extract the moving objects.

Finally we applied a data validation process by using a deep learning algorithm to classify the detected object as a vehicle. We trained a database consisted of 500 vehicle images by a convolutional neural network (CNN) [12]. Then we extracted the features of the candidate object by a CNN, after that we compared the features extracted of that object to the vehicles pre-trained CNN, then the best similarity is considered to find a match. To measure the similarity, two fully connected neural network layers are used.

## IV. SYSTEM RESULTS

Experiments have been performed to test the developed vehicles detection system with the four background modeling based on blocks methods, and to measure the accuracy of them. The systems are designed in MATLAB R2017, the test videos were taken under various illumination and mobile conditions, the results are as follows:

- First method which depends on absolute difference, achieved vehicles detection accuracy 82%
- Second method which depends on image entropy, achieved vehicles detection accuracy 89%
- Third method which depends on XOR function, achieved vehicles detection accuracy 93%
- Fourth method which depends on DCT, achieved vehicles detection accuracy 96%, which is the best one.

## V. CONCLUSIONS

We presented a complete developed system to detect vehicles in mobile and complex situations. The system based on background image modeling, which depending on segmenting image to equal size blocks; then applying four different methods (absolute difference, image entropy, XOR and DCT) to find the similarity between counterpart block pixels in the adjacent frames. After successfully background image modeling, we used a background image subtraction algorithm to detect moving objects by using XOR followed by median filter to obtain the foreground mask. Finally a data validation process used to classify the detected object as a vehicle, this process implemented by CNN deep learning algorithm. The experimental results show that the DCT method which used in background modeling has higher vehicle detection accuracy than other methods.